\pdfoutput=1
\documentclass[11pt]{article}
\usepackage[final]{acl} 
\usepackage{times}
\usepackage{latexsym}
\usepackage[T1]{fontenc}
\usepackage[utf8]{inputenc}
\usepackage{microtype}     
\usepackage{inconsolata}    
\usepackage{amsmath}
\usepackage{amssymb}
\usepackage{graphicx}
\usepackage{booktabs}       
\usepackage{multirow}
\usepackage{subcaption}     
\usepackage{tcolorbox}      

\usepackage{algorithm}
\usepackage{algorithmic}

\usepackage{enumitem}
\setlist[itemize]{nosep, leftmargin=*}
\setlist[enumerate]{nosep, leftmargin=*}

\usepackage{xcolor}
\newcommand{\wrong}[1]{\textcolor{red}{\textbf{#1}}}
\newcommand{\correct}[1]{\textcolor{teal}{\textbf{#1}}}

\usepackage{dblfloatfix} 
\title{CoRect: Context-Aware Logit Contrast for Hidden State Rectification to Resolve Knowledge Conflicts}
\author{
  Xuhua Ma$^{1}$, Richong Zhang$^{1*}$, Zhijie Nie$^{1}$ \\
  $^{1}$Beihang University, China \\
  \texttt{\{maxuhua25, zhangrc\}@act.buaa.edu.cn} \\
  $^{*}$Corresponding author
}
\begin{document}
\maketitle
\begin{abstract}
Retrieval-Augmented Generation (RAG) often struggles with \textbf{knowledge conflicts}, where model-internal parametric knowledge overrides retrieved evidence, leading to unfaithful outputs. Existing approaches are often limited, relying either on superficial decoding adjustments or weight editing that necessitates ground-truth targets. Through layer-wise analysis, we attribute this failure to a \textbf{parametric suppression} phenomenon: specifically, in deep layers, certain FFN layers overwrite context-sensitive representations with memorized priors. To address this, we propose \textsc{CoRect} (\textbf{Co}ntext-Aware Logit Contrast for Hidden State \textbf{Rect}ification).By contrasting logits from contextualized and non-contextualized forward passes, \textsc{CoRect} identifies layers that exhibit high parametric bias without requiring ground-truth labels. It then rectifies the hidden states to preserve evidence-grounded information. Across question answering (QA) and summarization benchmarks, \textsc{CoRect} consistently improves faithfulness and reduces hallucinations compared to strong baselines.
\end{abstract}

\section{Introduction}
\label{sec:intro}

\begin{figure}[t]
    \centering
    \includegraphics[width=1.0\columnwidth]{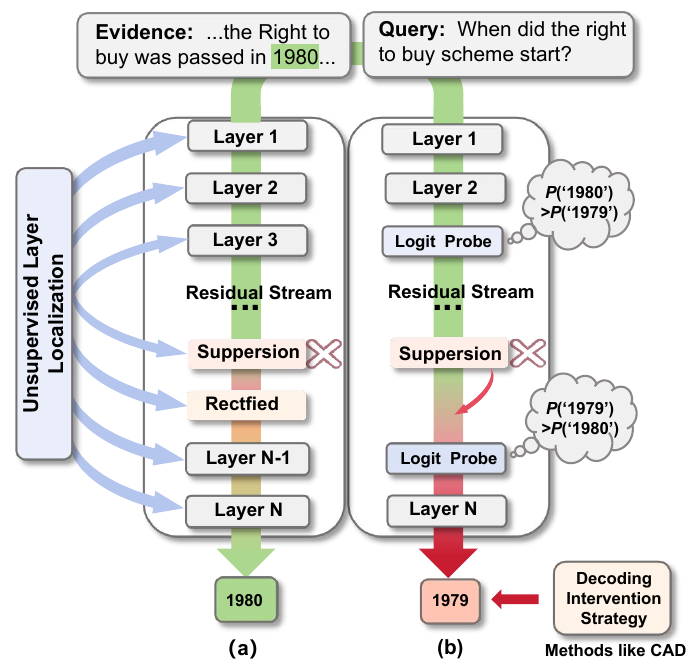}
    \caption{Comparison of intervention strategies. (a) Our method localizes and rectifies parametric suppression layers within the model's internal residual stream. (b) Baseline methods treat the model as a black box by intervening only at the output stage.} 
    \label{fig:intro_mechanism}
\end{figure}

Large Language Models (LLMs) have demonstrated remarkable capabilities in knowledge-intensive tasks~\cite{brown2020language, touvron2023llama}.However, because much of their knowledge is encoded in static pre-training corpora, LLMs often hallucinate or produce outdated statements when queried about evolving facts.Retrieval-Augmented Generation (RAG) mitigates this issue by conditioning generation on external evidence retrieved at inference time, thereby improving factuality and robustness~\cite{lewis2020retrieval, guu2020retrieval}.
Despite its promise, RAG often fails under \emph{knowledge conflict}---where the retrieved evidence contradicts the model's parametric prior knowledge~\cite{longpre2021entity, chen2022rich}. In such settings, LLMs tend to under-utilize the provided context, favoring internally memorized but incorrect knowleage. As illustrated in Figure~\ref{fig:intro_mechanism}, when queried about the start date of the right to buy scheme, LLMs favor internal biases 1979 even when context specifies the correct fact 1980.

Recent work has proposed decoding-time corrections to address this failure mode. Prominent approaches include adjusting the final token distribution by subtracting logits attributed to parametric priors~\cite{shi2024trusting, wang2025adacad}, or utilizing information entropy constraints to dynamically balance context and memory~\cite{yuan2024discerning}.
While effective in many scenarios, these methods largely treat the model as a \emph{black box} and intervene only at the output layer. Such late interventions are weakly grounded in the model's internal computation: they do not prevent intermediate representations from being progressively steered toward the parametric answer. Moreover, aggressive manipulation of the output distribution can disrupt the model's inherent linguistic priors, often leading to semantic instability or quality degradation.

In this work, we shift from output-level corrections to a mechanistic view of where knowledge conflicts emerge inside the model. Using a layer-wise analysis of the residual stream via Logit Lens~\cite{nostalgebraist2020logitlens}, we observe a consistent \emph{parametric suppression} phenomenon: under conflicts, although early layer representations initially reflect the retrieved evidence, deeper layers are progressively overwhelmed by internal priors, effectively suppressing the context to revert toward the and incorrect answer.Inspired by mechanistic interpretability and model editing studies, we hypothesize that this drift is driven by a subset of Feed-Forward Network (FFN) layers that act as key-value memories for parametric facts~\cite{geva2021transformer, dai2022knowledge}. Concretely, while attention mechanisms successfully incorporate external context, certain FFN layers overwrite this evidence-consistent state with parametric memories, thereby enforcing the prior over the context.

Once the failure is attributed to FFNs, a natural idea is to directly modify these modules. Indeed, ROME-style editing methods update weights in selected FFN layers to rewrite factual associations~\cite{meng2022locating, meng2022mass}. However, such approaches crucially assume access to the answer, which is unrealistic for open-ended generation in RAG. At inference time, the system typically does not know the single gold answer in advance. This limitation motivates a different goal: resolving conflicts without target answer during inference.

We therefore propose a training-free framework that mitigates knowledge conflicts via targeted hidden-state intervention. Our key idea is a dynamic layer localization strategy that identifies conflict-inducing layers where parametric suppression originates---without the correct answer. Once localized, we apply an inference-time intervention that blocks the propagation of conflict-induced parametric information and restores evidence-following behavior, while preserving the model's general generation capabilities.Our contributions are summarized as follows:
\begin{itemize} \item \textbf{Unveiling the Parametric Suppression Phenomenon.} We identify a phenomenon termed Parametric Suppression where deep FFN layers override representations with memorized priors. This insight isolates the specific internal components responsible for unfaithful generation.  \item \textbf{Target-Agnostic Inference-Time Intervention.} We propose a novel intervention framework that localizes conflict-inducing layers without target answers, achieving high alignment with over \textbf{70\%} recall relative to critical layers identified by ROME. By rectifying hidden states within the identified layers during inference, our approach effectively mitigates knowledge conflicts and reduces hallucinations while preserving fluency. \item \textbf{Effective Mitigation of Knowledge Conflicts} Extensive evaluations across Question Answering and Summarization tasks demonstrate the effectiveness of our framework, showing that it outperforms state-of-the-art baselines in resolving knowledge conflicts. \end{itemize}

\section{Related Work}
\label{sec:related_work}

We tackle RAG knowledge conflicts via decoding, model editing, and mechanistic interpretability.

\paragraph{Decoding-Time Interventions.}
Knowledge conflicts occur when models prioritize parametric priors over contradictory retrieved evidence~\cite{longpre2021entity, xie2023adaptive}. To mitigate this, strategies like CAD~\cite{shi2024trusting} and AdaCAD~\cite{wang2025adacad} adjust the final token distribution, typically by contrasting context-augmented logits against a context-free baseline.By treating the model as a black box, they fail to arrest the internal propagation of incorrect parametric information, potentially leading to generation instability despite maintaining surface-level fluency.

\paragraph{Mechanistic Localization and Editing.}
Going deeper than the output layer, mechanistic studies identify Feed-Forward Networks (FFNs) as key-value memories storing factual associations~\cite{geva2021transformer, dai2022knowledge}. Methods like ROME~\cite{meng2022locating} and MEMIT~\cite{meng2022mass} leverage this insight to permanently edit FFN weights. However, these approaches require the target answer to compute updates, rendering them unsuitable for open-ended RAG where the correct answer is unknown. Furthermore, permanent parameter modifications lack the flexibility to handle transient, query-specific conflicts.

\paragraph{Inference-Time Activation Intervention.}
Alternatively, activation engineering methods like ITI~\cite{li2024inference} and Representation Engineering~\cite{zou2023representation} steer model behavior by injecting static vectors derived from trained probes. While effective, they typically rely on labeled data for training and apply global, static interventions. In contrast, our framework is \textbf{fully training-free and instance-specific}. Instead of relying on static directions or supervised weight updates, we utilize the Logit Lens~\cite{nostalgebraist2020logitlens} to dynamically localize the specific layers where parametric suppression originates for each query, enabling targeted hidden-state intervention that resolves conflicts without pre-trained classifiers.
\section{Preliminaries}
\label{sec:prelim}
\subsection{Residual Decomposition for Transformer}
\label{ssec:residual_stream}
We analyze a standard $L$-layer decoder-only LLM, denoted as $F_{\mathrm{LLM}}$, with a vocabulary $\mathcal{V}$. In the prevalent pre-layer normalization (Pre-LN) architecture, the model structure can be formalized as:
\begin{equation}
   F_{\mathrm{LLM}} = W_{U} \circ f_{\mathrm{LN}} \circ f_L \circ \cdots \circ f_1 \circ W_E,
\end{equation}
where $W_{U} \in \mathbb{R}^{|\mathcal{V}|\times d}$ is the unembedding matrix, $W_{E} \in \mathbb{R}^{d\times |\mathcal{V}|}$ is the embedding matrix, and $f_{\mathrm{LN}}$ denotes layer normalization. The $l$-th Transformer layer $f_l$ comprises an attention module and a feed-forward network (FFN), formulated as:
\begin{equation}
    f_l = f^{(l)}_{\mathrm{ffn}} \circ f_{\mathrm{LN}} \circ f^{(l)}_{\mathrm{attn}} \circ f_{\mathrm{LN}},
\end{equation}
where $f^{(l)}_{\mathrm{ffn}}$ and $f^{(l)}_{\mathrm{attn}}$ denote the FFN and attention modules of the $l$-th layer, respectively. Let $h_{l} \in \mathbb{R}^d$ denote the residual stream state at layer $l$. The stream is updated additively:
\begin{equation}
    h_l = h_{l-1} + a_l + u_l,
\end{equation}
where $a_l = f^{(l)}_{\mathrm{attn}}(f_{\mathrm{LN}}(h_{l-1}))$ and $u_l = f^{(l)}_{\mathrm{ffn}}(f_{\mathrm{LN}}(h_{l-1} + a_l))$. The final logits $z_L$ are obtained by applying a decoding head to the normalized final state: $z_L = W_U f_{\mathrm{LN}}(h_L)$.

To analyze intermediate representations, we employ the Logit Lens technique~\cite{nostalgebraist2020logitlens}. By treating Layer Normalization as a locally linear operation or absorbing it into the projection, we can approximate the final logits as a linear decomposition of components from all layers:
\begin{equation}
\label{eq:logit_sum}
    z_L \approx W_U h_0 + \sum_{l=1}^{L} W_U a_l + \sum_{l=1}^{L} W_U u_l.
\end{equation}
Eq.~\eqref{eq:logit_sum} interprets the prediction as a superposition of the initial embedding and the outputs from attention heads and MLP blocks.
\subsection{Weight Editing as Residual Steering}
\label{ssec:editing_abstraction}

Most knowledge editing methods modify the parameters of Feed-Forward Network (FFN) modules, operating on the hypothesis that FFNs serve as key-value memories for factual knowledge~\cite{meng2022locating}. 
In the context of residual networks, the output of these modules, $u_l$, plays a critical role: it is added directly to the residual stream, thereby propagating the encoded factual information as a linear contribution to the final prediction.

\paragraph{General Framework.}
Given a prompt and a desired target token $t^*$, the objective of editing is to amplify the logit of $t^*$. Functionally, parameter edits can be abstracted as injecting a \textit{steering vector} $\delta_l$ into the residual pathway:
\begin{equation}
    \widehat{u}_l = u_l + \delta_l.
\end{equation}
Let $w_t$ represent the row vector of $W_U$ corresponding to token $t^*$. 
Adopting the view of the residual stream as a linear communication channel~\cite{elhage2021mathematical}, and following the decoding analysis in~\cite{geva2021transformer}, we approximate the cumulative effect of the edit as the projection of the steering vector onto the target's unembedding direction(A detailed derivation based on linear channel assumption is provided in Appendix~\ref{app:derivation}):
\begin{equation}
\label{eq:general_logit_shift}
    \Delta z_{L}(t^*) \approx \sum_{l \in \xi} w_{t^*}^\top \delta_l,
\end{equation}
where $\xi$ indexes the edited layers. Eq.~\eqref{eq:general_logit_shift} reveals that the optimal steering direction is explicitly aligned with $w_{t^*}$, making target-guided residual rectification inherently label-dependent.

\paragraph{Analysis of ROME.}
Formally, we define the \textit{post-activation key vector} $m_l$ as the intermediate activation produced by the up-projection:
\begin{equation}
\label{eq:ffn_decomposition}
    m_l = \sigma \!\left( W_{\mathrm{up}}^{(l)} f_{\mathrm{LN}}(h_{l-1} + a_l) \right).
\end{equation}
Consequently, the output of the FFN at layer $l$, which propagates to the residual stream, is computed as:
\begin{equation}
    u_l = W_{\mathrm{down}}^{(l)} m_l.
\end{equation}
where $\sigma$ is the activation function. 
ROME~\cite{meng2022locating} instantiates this abstraction by realizing $\delta_l$ via a rank-one weight update. ROME computes a parameter update $\Delta W^{(l)}$:
\begin{equation}
\label{eq:rome_implementation}
    \delta_l = \Delta W^{(l)} m_l.
\end{equation}
ROME(Detailed proofs are provided in Appendix \ref{app:rome_derivation}) constructs $\Delta W^{(l)}$ specifically to align the induced $\delta_l$ with the target direction $w_{t^*}$. This highlights a critical limitation: existing weight-editing approaches heavily rely on an \textit{oracle target} $t^*$ to determine the update direction. This dependency precludes their application in autonomous correction settings where the ground truth label is unavailable.

\section{Methodology}
\label{sec:method}
\begin{figure}[t]
    \centering
    \includegraphics[width=\linewidth]{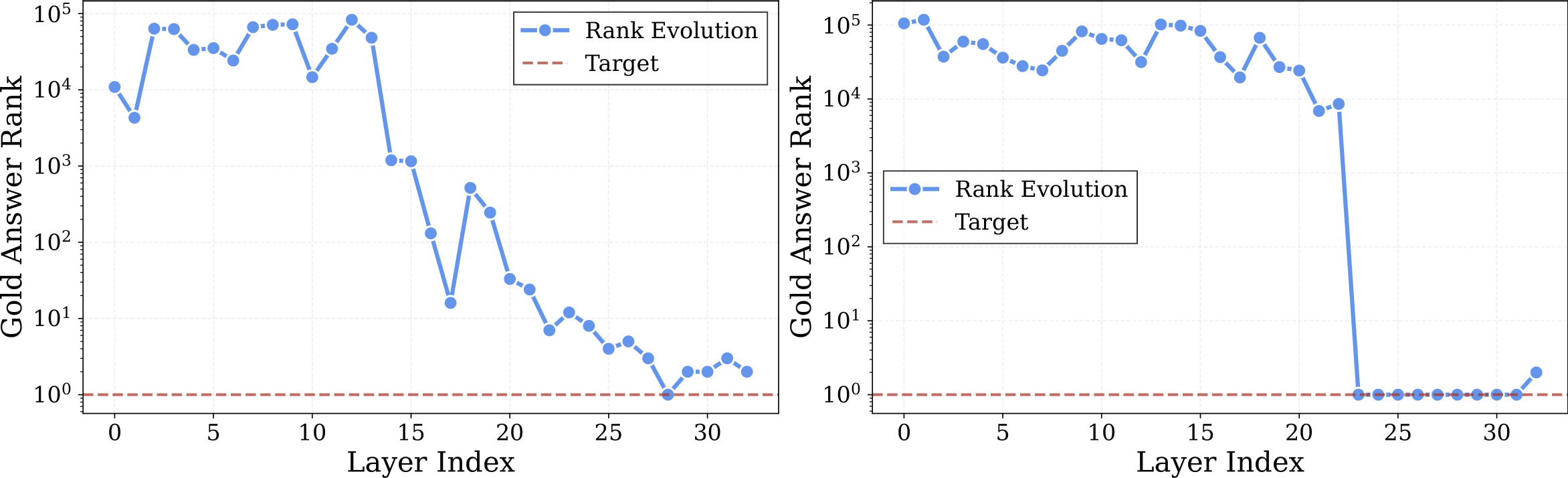}
    \caption{
        \textbf{Rank evolution analysis.} 
        (Top) Middle layer flip pattern. 
        (Bottom) Last layer flip pattern.
    }
    \label{fig:vertical_analysis}
\end{figure}
\begin{figure*}[t] 
    \centering 
    \includegraphics[width=\textwidth]{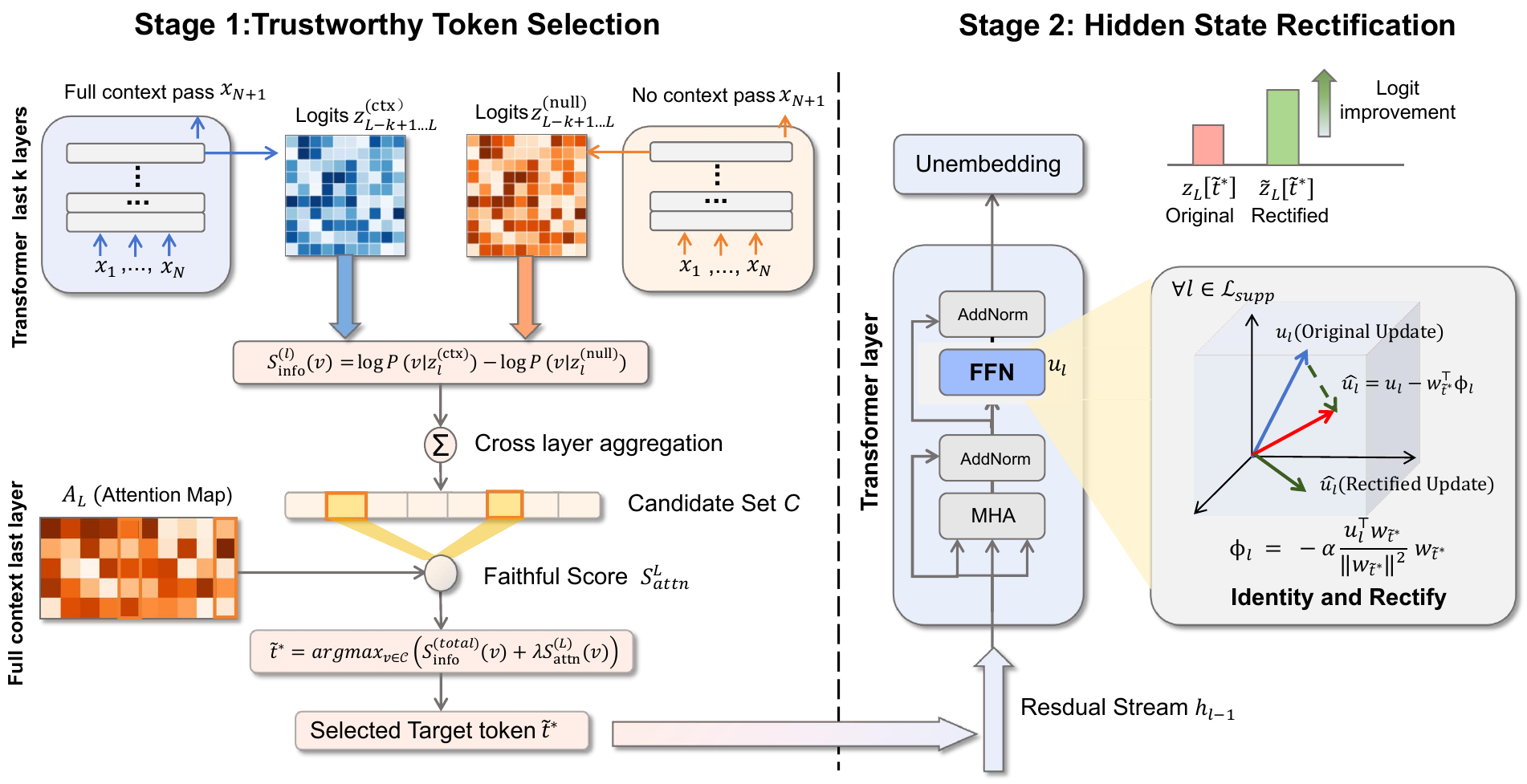} 
    \caption{The overall architecture of our proposed model.} 
    \label{fig:model}
\end{figure*}

\subsection{Parametric Suppression}
\label{ssec:parametric suppression}

To understand why the model fails to generate the correct answer despite potentially encoding it, we investigate the internal representation dynamics using the Logit Lens technique~\cite{nostalgebraist2020logitlens}. Formally, we project the hidden state $h_l$ of each layer $l$ directly onto the vocabulary space $\mathcal{V}$ using the pre-trained language modeling head $W_U$. This allows us to interpret the model's instantaneous prediction preference at specific depths of the network.We observe a counter-intuitive phenomenon where the gold answer token often achieves a high probability in intermediate layers, only to be suppressed in the final output. We term this phenomenon \textbf{Parametric Suppression}.

\paragraph{Qualitative Analysis.}
As illustrated in Figure~\ref{fig:vertical_analysis}, we analyze the rank evolution of the gold answer for the query: \textit{``How many episodes in season 5 of Curse of Oak Island?''}.
In the intermediate layers, the model's internal states strongly favor the correct answer 111, However, as the information propagates to the final layers, the rank of 111 drops significantly, and the model ultimately outputs the incorrect token 5.
This discrepancy suggests that the correct knowledge was successfully retrieved and encoded in the middle layers but was subsequently overridden during the forward pass.

To verify the prevalence of this phenomenon, we conducted a systematic evaluation on 500 sampled cases from the Natural Questions (NQ) training set that involve knowledge conflicts. We tracked the rank evolution of the gold answers across all layers.
Our analysis reveals that \textbf{282 out of 500 cases} exhibit this flip phenomenon: the gold answer reaches Rank 1 at some point but is not the final prediction.
Specifically, we categorize these into two patterns:
Middle Flip (154 cases):The gold answer surfaces in the intermediate layers but is lost in the later stages.
Last Layer Flip (128 cases):The gold answer remains dominant until the very last projection layer, where it is suddenly suppressed.The high prevalence of parametric suppression indicates that the model often possesses the  capability to solve the problem, yet fails to manifest it in the final generation.
While we acknowledge that some rank fluctuations may reflect legitimate reasoning steps, the fact that over 56\% of error cases contain the correct answer internally serves as a strong motivation for our approach.
It suggests that instead of injecting new knowledge, we can improve performance by intervening in the decoding process to release the suppressed correct information.
\subsection{Transition: From Supervised Overwriting to Autonomous Rectification}
\label{ssec:transition}
The foregoing analysis underscores that conventional editors operationalize editing as \emph{supervised overwriting}: an externally specified $t^*$ is used to construct a parameter patch that compels the model toward a designated output. In contrast, our approach departs from this paradigm along two axes---\emph{how the target is obtained} and \emph{how the intervention is applied}.
we introduce a two-stage procedure:
(1) \textbf{Token Selection} We infer a reliable target token $\tilde{t}^*$ from the input prompt itself, thereby removing the requirement for oracle target;
(2) \textbf{Hidden State Rectification} Instead of forcefully steering generation toward $\tilde{t}^*$, we apply a minimal intervention to the hidden representation $u_l$ that selectively attenuates components that \emph{inhibit} $\tilde{t}^*$. Conceptually, we frame this as disinhibition rather than injection. By removing suppressive components, we aim to preserve the model's native generative dynamics while selectively mitigating hallucinations.The overall architecture of our proposed model is Figure~\ref{fig:model}.

\subsection{Stage 1: Trustworthy Token Selection}
\label{ssec:token_selection}
Simple subtraction of logits often amplifies noise alongside the correct answer. To select a robust target token $\tilde{t}^*$ at current step, we integrate Contextual Mutual Information with attention-based evidence.
\paragraph{Contextual Mutual Information Estimation}
We perform two forward passes: one with the full input prompt and one with a null context consisting only of the instruction tokens, so that the null pass captures generic prior preferences independent of the specific query.
Let $z_l^{(\mathrm{ctx})}$ and $z_l^{(\mathrm{null})}$ denote the logits at layer $l$, obtained by projecting the layer's hidden state onto the vocabulary using the logit lens.
For each token $v \in \mathcal{V}$, we compute the layer-wise $S_{\mathrm{info}}^{(l)}(v)$ which captures the boost in likelihood for token $v$ provided by the context: 
\begin{equation}
    S_{\mathrm{info}}^{(l)}(v) = \log P(v \mid z_l^{(\mathrm{ctx})}) - \log P(v \mid z_l^{(\mathrm{null})}),
\end{equation}
where $P(\cdot \mid z) = \mathrm{softmax}(z)$ represents the probability distribution over the vocabulary given logits $z$. 
To capture the evolution of information, we focus on the last $k$ layers. We aggregate these scores via the mean:
\begin{equation}
    \tilde{S}_{\mathrm{info}}(v) = \frac{1}{k}\sum_{l = L-k+1}^L S_{\mathrm{info}}^{(l)}(v).
\end{equation}
To ensure numerical stability, we apply sign-preserving max-normalization. The global information score is defined as:
\begin{equation}
\label{eq:score_norm}
    S_{\mathrm{info}}^{(\text{total})}(v) = \frac{\tilde{S}_{\mathrm{info}}(v)}{\max_{v' \in \mathcal{V}} \left| \tilde{S}_{\mathrm{info}}(v') \right| + \epsilon}.
\end{equation}
where $\epsilon$ is a small scalar for numerical stability.
\paragraph{Candidate Selection \& Attention Filter.}
We define the candidate set $\mathcal{C}$ as the top-$M$ tokens ranked by $S_{\mathrm{info}}^{(\text{total})}$.
Then we employ an attention-based verification mechanism that favors candidates grounded in the input context.
For each candidate $v \in \mathcal{C}$, we perform a token matching step to identify its occurrences in the context $\tilde{\mathbf{x}}$ of the input $\mathbf{x}$. Let $\mathcal{I}(v) = \{ j \mid \tilde{x}_j = v \}$ denote the set of position indices where the token $v$ appears.
We then quantify the evidence for $v$ by aggregating the attention mass the model allocates to these matched positions:
\begin{equation}
    \tilde{S}_{\mathrm{attn}}^{(L)}(v) = 
    \begin{cases} 
    \displaystyle \frac{1}{H}\sum_{h=1}^{H}\sum_{j\in \mathcal{I}(v)} A_{L,h}^{(\mathrm{curr},j)}, & \text{if } \mathcal{I}(v) \neq \emptyset \\
    0, & \text{otherwise}.
    \end{cases}
\end{equation}
where $H$ represents the total number of attention heads. The term $A_{L,h}^{(\mathrm{curr},j)}$ denotes the attention weight assigned by the $h$-th head at layer $L$, This formulation essentially averages the attention mass that the model's final layer directs toward the candidate $v$.This metric serves as a \textit{Faithfulness Score}, ensuring that the selected target is not only informative but also supported by the model's internal attention to the context.
We apply the same sign-preserving max-normalization (as defined in Eq.~\ref{eq:score_norm}) to the raw score $\tilde{S}_{\mathrm{attn}}^{(L)}(v)$ to obtain $S_{\mathrm{attn}}^{(L)}(v)$.
At current autoregressive decoding step $t$, the step-wise target token $\tilde{t}_t^*$ is selected to maximize the following joint objective:
\begin{equation}
\label{eq:final_target}
    \tilde{t}^* = \operatorname*{argmax}_{v \in \mathcal{C}} \left( S_{\mathrm{info}}^{(\text{total})}(v) + \lambda S_{\mathrm{attn}}^{(L)}(v) \right),
\end{equation}
where $\lambda$ balances contextual induction with explicit evidence support, thereby offering the flexibility to accommodate varying demands ranging from extractive grounding to logical deduction.

\subsection{Stage 2: Hidden State Rectification}
\label{ssec:rectification}

Given the target unembedding vector $w_{\tilde{t}^*}$, we identify the \emph{suppressive layers} $\mathcal{L}_{supp}$ where the MLP's direct contribution negatively impacts the target:
\begin{equation}
\label{eq:suppressive_def}
    \mathcal{L}_{supp} = \left\{ l \in \{1, \dots, L\} \;\middle|\; u_l^\top w_{\tilde{t}^*} < 0 \right\}.
\end{equation}
To mitigate this, we construct an additive patch $\phi_l$ to cancel the projection of $u_l$ onto the $w_{\tilde{t}^*}$ direction.
We define the rectification patch:
\begin{equation}
\label{eq:rectification_patch}
\begin{split}
    \phi_l \;&=\; -\alpha \frac{u_l^\top w_{\tilde{t}^*}}{\|w_{\tilde{t}^*}\|^2} \, w_{\tilde{t}^*}, \quad \forall l \in \mathcal{L}_{supp}.
\end{split}
\end{equation}
This formulation offers a tunable spectrum of intervention controlled by $\alpha$. 
Specifically, setting $\alpha=1$ reduces the term to zero, strictly \textit{neutralizing} the suppression. This corresponds to the pure ``disinhibition'' paradigm discussed in Sec.~\ref{ssec:transition}, where the blockage is removed without imposing new directional force. 
In contrast, setting $\alpha > 1$ flips the sign of the contribution from negative to positive, transitioning the operation from mere removal of inhibition to active \textit{promotion}of evidence for $\tilde{t}^*$.
Regardless of the specific regime, the net change in the final logit is geometrically guaranteed to be positive:
\begin{equation}
    \Delta z_L(\tilde{t}^*) \triangleq \sum_{l \in \mathcal{L}_{supp}} w_{\tilde{t}^*}^\top \phi_l = -\alpha \sum_{l \in \mathcal{L}_{supp}} w_{\tilde{t}^*}^\top u_l > 0.
\end{equation}
To verify the optimality of our identified layer $l$ compared to ROME's $l^\star$(the derivation of which is detailed in Appendix~\ref{app:rome_localization_final}), we analyze the editing performance across different layers.
\begin{figure}[t]
  \centering
  \includegraphics[width=\linewidth]{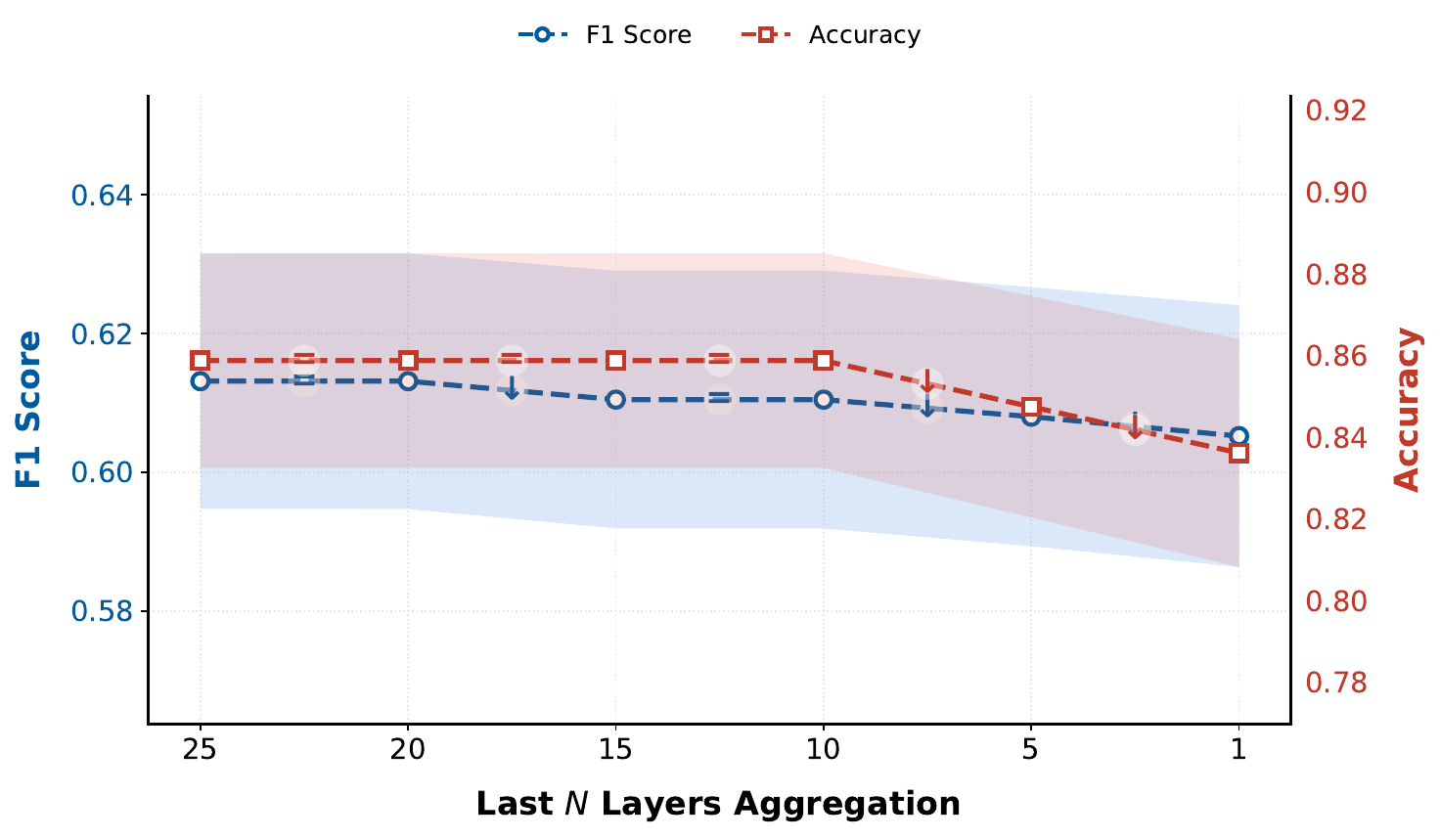}
  \caption{Performance metrics across different layer aggregation scales. The F1 score (blue) validates the localization accuracy by comparing ROME-identified layers $l^*$ with our defined layers $l$, while the accuracy (red) illustrates the performance stability after applying our correction method.}
  \label{fig:q1_performance}
\end{figure}
Figure~\ref{fig:q1_performance} illustrates the trends of F1 score and QA accuracy as we vary the intervention layer. 
\section{Experiments}
\label{sec:exp}
\begin{table*}[t]
\centering
\small
\setlength{\tabcolsep}{3.5pt}
\resizebox{\textwidth}{!}{
\begin{tabular}{llccccccc}
\toprule
\textbf{Model} & \textbf{Method} & \textbf{NQ} & \textbf{NQ-Swap} & \textbf{HotpotQA} & \textbf{TriviaQA} & \textbf{TabMWP} & \textbf{SQuAD} & \textbf{Avg} \\
\midrule
\multirow{5}{*}{\textbf{Qwen2.5-7b-instruct}} 
 & Greedy & 69.49 & 70.54 & 43.30 & 76.85 & 59.90 & 81.27 & 66.89 \\
 & CAD & 67.28 & 79.37 & 40.31 & 76.75 & 58.90 & 84.43 & 67.84 \\
 & COIECD & 59.87 & 78.56 & 41.90 & \textbf{80.75} & 64.35 & 88.90 & 69.06 \\
 & AdaCAD & 68.54 & 73.14 & 43.82 & 76.90 & 60.00 & 81.20 & 67.27 \\
 & \textbf{CoRect} & \textbf{72.74} & \textbf{80.15} & \textbf{45.67} & 79.60 & \textbf{70.60} & \textbf{88.93} & \textbf{72.95} \\
\midrule
\multirow{5}{*}{\textbf{LLaMA3-8b-instruct}} 
 & Greedy & 68.72 & 60.67 & 39.47 & 80.30 & 47.40 & 81.93 & 63.08 \\
 & CAD & 68.17 & \textbf{80.10} & 36.44 & 81.20 & 48.90 & \textbf{84.93} & 66.62 \\
 & COIECD & 60.23 & 70.39 & 37.10 & 80.95 & 51.95 & 84.40 & 64.17 \\
 & AdaCAD & 67.57 & 67.37 & 40.43 & 80.30 & 47.30 & 81.87 & 64.14 \\
 & \textbf{CoRect} & \textbf{71.22} & 79.32 & \textbf{41.15} & \textbf{83.00} & \textbf{52.60} & 83.70 & \textbf{68.50} \\
\midrule
\multirow{5}{*}{\textbf{Mistral-7b}} 
 & Greedy & 46.34 & 44.51 & 31.60 & 72.10 & 25.20 & 63.93 & 47.28 \\
 & CAD & 33.50 & 65.72 & 29.17 & 72.50 & 35.40 & 72.90 & 51.53 \\
 & COIECD & 36.85 & 42.96 & 29.23 & 76.35 & 36.50 & 74.20 & 49.35 \\
 & AdaCAD & 44.56 & 63.99 & 34.69 & 72.00 & 25.40 & 63.87 & 50.75 \\
 & \textbf{CoRect} & \textbf{56.31} & \textbf{70.37} & \textbf{36.06} & \textbf{78.60} & \textbf{38.50} & \textbf{74.33} & \textbf{59.03} \\
\midrule
\multirow{5}{*}{\textbf{LLaMA2-13b-chat}} 
 & Greedy & 44.15 & 49.29 & 21.27 & 55.60 & 17.90 & 52.17 & 40.06 \\
 & CAD & 46.88 & 74.59 & 23.53 & 63.55 & \textbf{27.30} & 68.97 & 50.80 \\
 & COIECD & 43.11 & 54.41 & 20.83 & 60.35 & 26.70 & 58.67 & 44.01 \\
 & AdaCAD & 43.40 & 57.69 & 29.13 & 55.05 & 17.70 & 51.80 & 42.46 \\
 & \textbf{CoRect} & \textbf{49.01} & \textbf{76.52} & \textbf{29.45} & \textbf{68.15} & 27.20 & \textbf{70.70} & \textbf{53.51} \\
\bottomrule
\end{tabular}
}
\caption{Comparison of different methods across multiple models on NQ, NQ-Swap, HotpotQA, TriviaQA,TabMWP and SQuAD. }
\label{tab:main_results}
\end{table*}

\begin{table*}[t]
\centering
\small
\setlength{\tabcolsep}{6pt}
\begin{tabular}{lccc|ccc|cc} 
\toprule
\multirow{2}{*}{\textbf{Method}} & \multicolumn{3}{c}{\textbf{CNN-DailyMail}} & \multicolumn{3}{c}{\textbf{XSum}} & \multicolumn{2}{c}{\textbf{TofuEval (AlignScore)}} \\
\cmidrule(lr){2-4} \cmidrule(lr){5-7} \cmidrule(lr){8-9}
 & \textbf{ROUGE-L} & \textbf{BERT-P} & \textbf{AlignScore} & \textbf{ROUGE-L} & \textbf{BERT-P} & \textbf{AlignScore} & \textbf{Overall} & \textbf{Main / Marginal} \\
\midrule
Greedy & 21.08 & 85.83 & 91.48 & 16.42 & 86.56 & 77.65 & 43.65 & 49.92 / 37.39 \\
CAD    & 18.40 & 84.66 & 76.91 & 15.25 & 84.30 & 67.00 & 50.33 & 53.42 / 47.24 \\
COIECD & 21.17 & 85.86 & 91.63 & 15.77 & 86.48 & 81.06 & 49.61 & 64.26 / 34.95 \\
AdaCAD & 21.27 & 85.63 & 91.87 & 15.81 & 86.43 & 82.02 & 57.11 & 66.33 / 47.89 \\
\midrule
\textbf{CoRect} & \textbf{21.97} & \textbf{86.03} & \textbf{92.88} & \textbf{20.04} & \textbf{87.30} & \textbf{83.30} & \textbf{69.45} & \textbf{73.69 / 65.21} \\
\bottomrule
\end{tabular}
\caption{Abstractive summarization performance on LLaMA-3-8b. CoRect significantly improves factual alignment (AlignScore) while maintaining generation quality.}
\label{tab:summarization}
\end{table*}

We evaluate our proposed method on both Question Answering (QA) and Summarization tasks. Our experiments aim to demonstrate the effectiveness of our method in mitigating hallucinations and prioritizing contextual faithfulness compared to strong inference-time intervention baselines.Detailed formalizations of the decoding-time correction baselines and the specific hyperparameter configurations for all methods are provided in Appendix~\ref{sec:decoding_methods} and Appendix~\ref{appendix:implementation_details}, respectively.
\subsection{Experimental Setup}
\paragraph{Datasets and Metrics.}
To ensure a comprehensive evaluation, we test CoRect across a wide spectrum of tasks. For \textbf{QA and Reasoning}, we cover NQ~\cite{kwiatkowski2019natural}, HotpotQA~\cite{yang2018hotpotqa}, TabMWP~\cite{lu2022dynamic}, TriviaQA~\cite{joshi2017triviaqa}, SQuAD~\cite{rajpurkar2016squad}and standard/high-conflict benchmarks NQ-Swap~\cite{longpre2021entity}. For \textbf{Summarization}, we include short/long-form synthesis XSum~\cite{narayan2018don}, CNN-DM~\cite{hermann2015teaching}) and topic-aligned evaluation (TofuEval~\cite{tang2024tofueval}. We report Exact Match (EM) for QA, alongside ROUGE-L~\cite{lin2004rouge}, BERTScore-P~\cite{zhang2019bertscore}, and AlignScore~\cite{zha2023alignscore} to rigorously assess faithfulness. Detailed dataset statistics and metric formalizations are provided in Appendix~\ref{app:datasets}.
\begin{figure*}[t]
    \centering
    \includegraphics[width=\textwidth]{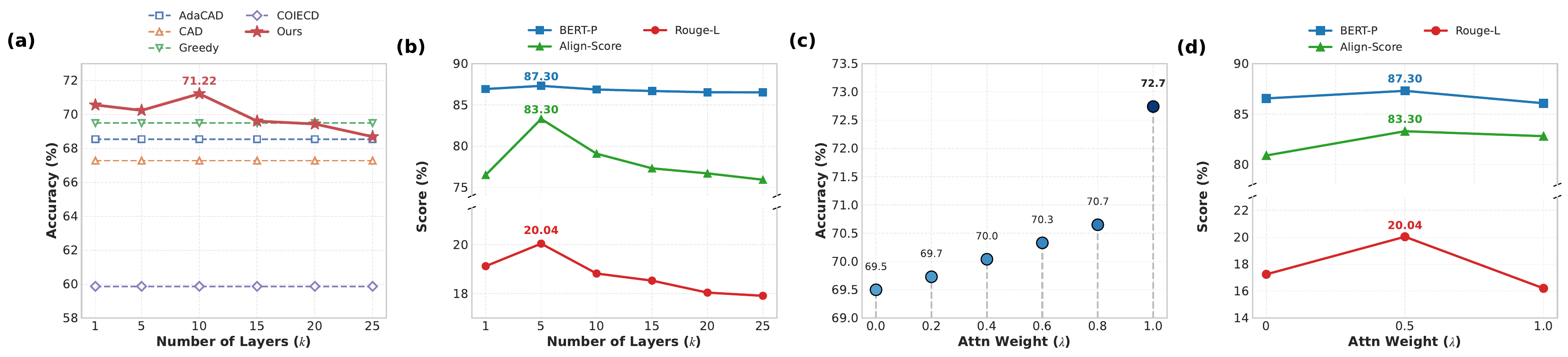} 
    \caption{
        \textbf{Hyperparameter sensitivity and performance analysis.} 
        \textbf{(a)} Accuracy with varying number of layers $K$ on NQ. 
        \textbf{(b)} Effect of layer depth on generation scores on XSUM. 
        \textbf{(c)} Impact of attention weight ($\lambda$) on model accuracy on NQ. 
        \textbf{(d)} Impact of attention weight ($\lambda$) on generation scores on XSUM.
    }
    \label{fig:hyperparameter_analysis}
\end{figure*}
\subsection{Main Results}
\paragraph{Question Answering.}
Table~\ref{tab:main_results} summarizes the performance across QA benchmarks. CoRect achieves consistently strong EM scores on TriviaQA and SQuAD, while demonstrating superior evidence extraction on NQ, even within long-context documents. Most notably, on NQ-Swap—a high-conflict stress test—CoRect yields substantial gains, showcasing its robustness against direct contradictions between context and priors. Furthermore, CoRect consistently improves performance on reasoning-intensive tasks like HotpotQA and TabMWP across various backbones.
\paragraph{Summarization and Alignment.}
We evaluate CoRect on summarization benchmarks (Table~\ref{tab:summarization}). CoRect consistently outperforms strong baselines across XSum, CNN/DailyMail, and TofuEval, demonstrating superior generalization across varying summary lengths and topic constraints.

\subsection{Ablation Studies}
\label{sec:ablation}
\paragraph{Q1: How Effective is the Token Selection Strategy?}
We compared our token selection strategy against baseline methods. As shown in Figure~\ref{fig:hyperparameter_analysis}(a), when the number of intervened layers $k$=1, our token selection strategy begins to above the baseline method, In this setting, our method demonstrates competitive performance on both the \textbf{NQ} and \textbf{XSum} datasets, confirming the effectiveness of our token selection strategy. 
\paragraph{Q2: Does the number of intervened layers $k$ matter?}
We analyzed the performance on NQ by varying the number of last layers \( k \in \{1, 5, 10, 15, 20, 25\} \). As shown in Figure~\ref{fig:hyperparameter_analysis}(a), performance peaks at \( k = 10 \), but slightly degrades as we intervene with too many layers. Additionally, we conducted similar experiments on the XSum dataset. As shown in Figure~\ref{fig:hyperparameter_analysis}(b), performance was maximized at  k = 5, where the summary quality metrics reached their highest values. These findings suggest that, while increasing the number of layers can help capture more nuanced information, excessive layer interventions may lead to the inclusion of irrelevant or noisy tokens, negatively affecting model performance on both NQ and XSum.
\paragraph{Q3: Is the Attention Filter Useful?}We applied the Attention Filter component on both the NQ and XSum datasets. As shown in Figure~\ref{fig:hyperparameter_analysis}(c)(d), the results demonstrate that the attention weight, denoted by $\lambda$, plays a crucial role in performance optimization. On the \textbf{NQ} dataset, performance improves gradually as the attention weight increases, reflecting the benefit of utilizing more context into the model’s responses. On the \textbf{XSum} dataset, the performance peaked at $\lambda = 0.5$. This difference can be attributed to the nature of the task. Unlike QA, where answers are typically extracted directly from the context , the summarization task in XSum requires the model to generate responses based on a broader understanding of the content. High attention weights in this case may cause the model to focus too heavily on specific parts of the input, which could limit its ability to generate a coherent and high-quality summary. Therefore, a moderate attention weight of $\lambda = 0.5$ strikes the optimal balance between context understanding and generating summaries with appropriate semantic depth.
\paragraph{Q4: Is Hidden State Rectification Effective?}
\begin{figure}[t]
  \centering
  \includegraphics[width=\linewidth]{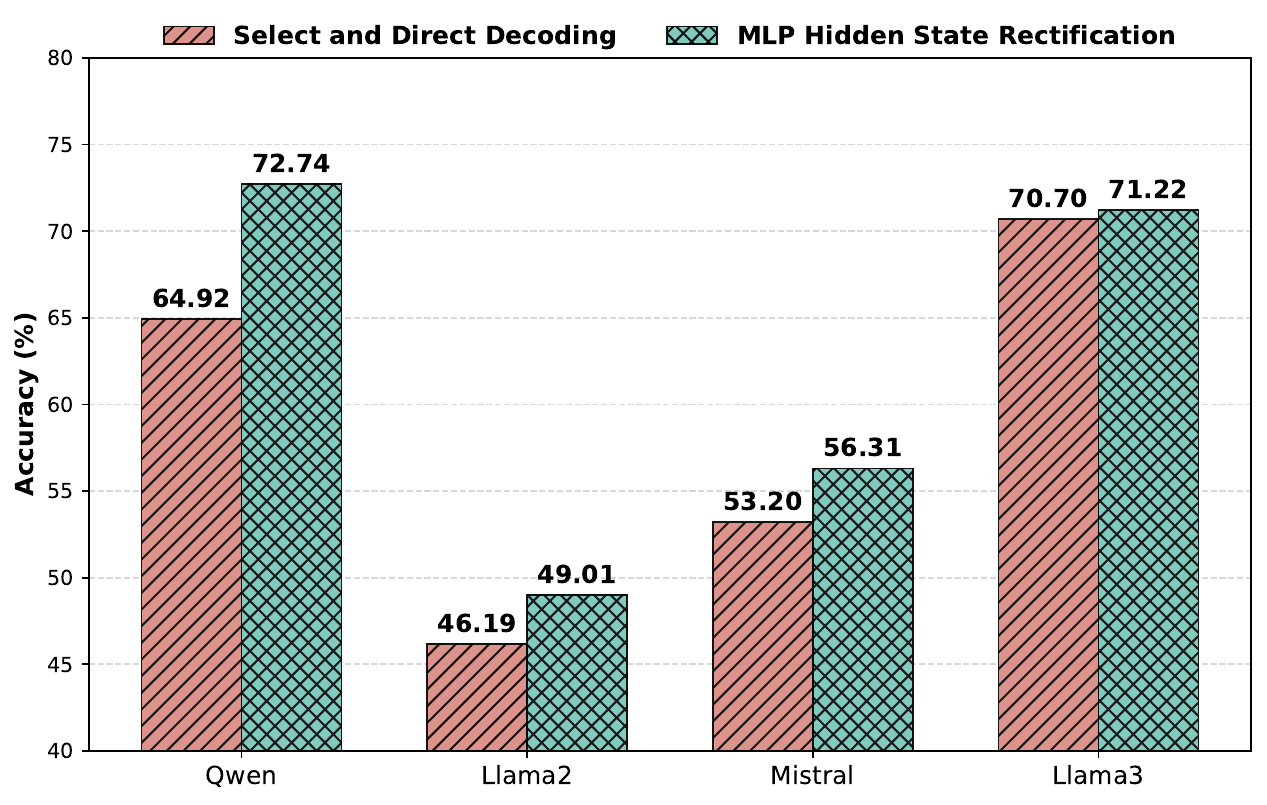}
  \caption{Accuracy comparison of different decoding strategies. Our Hidden State Rectification consistently outperforms Direct Decoding across all backbones.}
  \label{fig:q4_decoding}
\end{figure}
\begin{figure}[ht]
    \centering
    \footnotesize 
    \begin{tcolorbox}[
        width=\linewidth,      
        colback=white, 
        colframe=black, 
        boxrule=0.6pt,         
        arc=0pt, 
        outer arc=0pt,         
        top=3pt, 
        bottom=3pt, 
        left=4pt, 
        right=4pt,
        boxsep=0pt,            
        sharp corners,         
        nobeforeafter         
    ]
        \textbf{Context:} Suddenly is a song from Xanadu... performed as a duet between Olivia Newton-John and Cliff Richard...
        
        \vspace{3pt}
        \hrule height 0.5pt   
        \vspace{5pt}

        \textbf{Q:} who sang the song suddenly with olivia newton john? \hfill \textbf{Gold:} Cliff Richard \\
        \textbf{Direct Decoding:} \wrong{Ciff} Richard sang the song... \\
        \textbf{Ours (Rect.):} \correct{Cliff} Richard sang the song...
    \end{tcolorbox}
    \vspace{-8pt} 
    \caption{Our method rectifies the hidden state to generate the (\correct{Cliff}) where direct decoding fails (\wrong{Ciff}).}
    \label{fig:case_study_rectification}
\end{figure}
To evaluate the effectiveness of the hidden state rectification component, we compare the performance of directly decoding the target token selected in Stage 1 with that of using hidden state rectification. As shown in Figure ~\ref{fig:q4_decoding}, our method outperforms the direct decoding baseline. The core reason for this improvement lies in the fact that while the target token $\tilde{t}^*$ selected in Stage 1 provides a reliable direction, it is not always the optimal final token4. Direct decoding forces the model toward $\tilde{t}^*$, which can sometimes lead to semantic incoherence. In contrast, our rectification mechanism treats the selected token as a guiding vector to neutralize internal parametric suppression. This approach allows the model to maintain its native generative capabilities while surfacing the correct information, resulting in higher accuracy and better linguistic quality.A representative case demonstrating the efficacy of our approach is illustrated in Figure ~\ref{fig:case_study_rectification}.

\section{Conclusion}
\label{sec:conclusion}
In this work, we unveil Parametric Suppression, a phenomenon where deep FFN layers in LLMs override retrieved evidence and revert to memorized priors under knowledge conflicts. To address this, we propose CoRect, a target-agnostic hidden-state rectification method that localizes conflict-inducing layers and applies targeted interventions to neutralize parametric bias. Extensive evaluations demonstrate that improving RAG faithfulness is more effectively achieved via internal mechanistic corrections rather than via superficial output-level calibration. Ultimately, CoRect offers a practical, interpretable and training-free framework for solving knowledge conflicts in real-world.
\section{Limitations}
While CoRect effectively mitigates knowledge conflicts, it introduces non-negligible computational overhead compared to standard decoding. The framework necessitates a multi-pass inference scheme and stage-wise hidden state rectification, which entails frequent access to residual streams across layers. To alleviate this temporal burden, KV cache can be reused across passes to minimize redundant prefix computations, preventing a strictly linear growth in latency. While CoRect still introduces additional latency compared to vanilla decoding, this trade-off facilitates a more mechanistically interpretable and context-faithful resolution of knowledge conflicts.
\bibliography{custom}

\appendix  
\section{Derivation of the Optimal Steering Direction}
\label{app:derivation}

In this section, we provide a formal justification for Eq.~\eqref{eq:general_logit_shift} and the optimality of aligning $\delta_l$ with $w_{t^*}$.

\subsection{First-Order Approximation of Logit Shift}
Let $h_L$ be the final hidden state of the Transformer before the unembedding layer. The logit for the target token $t^*$ is given by the inner product $z_L(t^*) = w_{t^*}^\top h_L$. When we introduce perturbations $\{\delta_l\}_{l \in l^*}$ at intermediate layers, the change in the target logit $\Delta z_L(t^*)$ can be estimated using a first-order Taylor expansion:
\begin{equation}
    \Delta z_L(t^*) \approx \sum_{l \in l^*} \left( \nabla_{h_l} z_L(t^*) \right)^\top \delta_l
\end{equation}
where $h_l$ is the hidden state at layer $l$. Applying the chain rule, the gradient with respect to $h_l$ is:
\begin{equation}
    \nabla_{h_l} z_L(t^*) = \left( \frac{\partial h_L}{\partial h_l} \right)^\top \nabla_{h_L} z_L(t^*) = \mathbf{J}_{l \to L}^\top w_{t^*}
\end{equation}
where $\mathbf{J}_{l \to L} = \frac{\partial h_L}{\partial h_l}$ is the Jacobian matrix representing the transformation of the residual stream through subsequent layers.

\subsection{Linear Communication Channel Assumption}
Following \cite{elhage2021mathematical}, the linear communication channel hypothesis assumes that the residual stream acts as a high-fidelity conveyor where information is primarily transmitted linearly. Mathematically, this implies that for the relevant steering components, the Jacobian matrix is approximately identity:
\begin{equation}
    \mathbf{J}_{l \to L} \approx \mathbf{I}
\end{equation}
Substituting this into the expansion, we obtain the approximation used in the main text:
\begin{equation}
    \Delta z_L(t^*) \approx \sum_{l \in l^*} w_{t^*}^\top \delta_l
\end{equation}

\subsection{Proof of Optimality}
To find the optimal steering direction $\delta_l$, we consider the optimization problem under a fixed norm constraint $\|\delta_l\| = \epsilon$ for each layer:
\begin{equation}
    \max_{\delta_l} \quad w_{t^*}^\top \delta_l \quad \text{s.t.} \quad \|\delta_l\| = \epsilon
\end{equation}
By the Cauchy-Schwarz inequality:
\begin{equation}
    |w_{t^*}^\top \delta_l| \leq \|w_{t^*}\| \|\delta_l\|
\end{equation}
The equality (and thus the maximum value) is achieved if and only if $\delta_l$ is linearly dependent on $w_{t^*}$ with a positive scalar:
\begin{equation}
    \delta_l \propto w_{t^*}
\end{equation}
This confirms that under the linear channel assumption, the optimal direction to rectify the residual stream for a target token $t^*$ is to align the perturbation with its corresponding unembedding vector $w_{t^*}$.

\section{Derivation of Rank-One Model Editing}
\label{app:rome_derivation}

This appendix details the derivation of the ROME update rule referenced in Section~\ref{ssec:editing_abstraction}. We explicitly show how the optimization objective translates into the rank-one weight update and connect this back to the residual steering framework introduced in Eq.~\eqref{eq:rome_implementation}.

\subsection{Constrained Optimization Problem}
Let $W_0$ be the original weight matrix of the FFN down-projection at layer $l$. In the notation of the main text, the input to this layer is the post-activation vector $m_*$ (associated with the subject token), and the original output is $u_* = W_0 m_*$.

ROME seeks an updated matrix $\widehat{W}$ that satisfies two conflicting objectives:
\begin{enumerate}
    \item \textbf{Target Alignment (The Edit):} For the specific subject key $m_*$, the output should map to a new target vector $v_*$. This $v_*$ is optimized specifically to maximize the probability of the target token $t^*$:
    \begin{equation}
        \widehat{W} m_* = v_*
    \end{equation}
    \item \textbf{Knowledge Preservation:} For all other keys $m$ drawn from the general distribution of text (characterized by covariance $C = \mathbb{E}[mm^T]$), the output should remain unchanged:
    \begin{equation}
        \min_{\widehat{W}} \mathbb{E}_{m \sim C} [ \|\widehat{W}m - W_0 m\|^2 ]
    \end{equation}
\end{enumerate}

Defining $\Delta W = \widehat{W} - W_0$, the problem simplifies to minimizing the norm of the change weighted by the covariance $C$, subject to the linear constraint:
\begin{equation}
    \begin{aligned}
        & \min_{\Delta W} \quad \text{Tr}(\Delta W C \Delta W^T) \\
        & \text{s.t.} \quad \Delta W m_* = v_* - W_0 m_*
    \end{aligned}
\end{equation}

\subsection{Analytical Solution}
Using Lagrange multipliers (similar to the standard derivation in~\cite{meng2022locating}), the optimal closed-form solution is given by:
\begin{equation}
\label{eq:rome_update_formula}
    \Delta W = \frac{v_* - W_0 m_*}{m_*^T C^{-1} m_*} (C^{-1} m_*)^T
\end{equation}
This confirms that the update is a rank-one matrix, defined by the outer product of the \textit{residual vector} $(v_* - W_0 m_*)$ and the \textit{whitened key direction} $(C^{-1} m_*)^T$.

\subsection{Connection to Residual Steering and Target Dependency}
\label{app:connection_steering}
We now link this result back to the residual steering abstraction in Section~\ref{ssec:editing_abstraction}. 
Substituting the derived update $\Delta W$ (Eq.~\ref{eq:rome_update_formula}) into the steering equation (Eq.~\ref{eq:rome_implementation}), the effective steering vector $\delta_l$ injected into the residual stream for the subject $m_*$ is:

\begin{equation}
    \begin{aligned}
        \delta_l &= \Delta W m_* \\
        &= \left( \frac{v_* - W_0 m_*}{m_*^T C^{-1} m_*} (C^{-1} m_*)^T \right) m_* \\
        &= (v_* - W_0 m_*) \frac{m_*^T C^{-1} m_*}{m_*^T C^{-1} m_*} \\
        &= v_* - W_0 m_*
    \end{aligned}
\end{equation}

This derivation explicitly reveals the dependency discussed in the main text:
\begin{itemize}
    \item The steering vector $\delta_l = v_* - u_{original}$ is exactly the difference required to shift the representation from the old knowledge to the new target vector $v_*$.
    \item Since $v_*$ is optimized to maximize $P(t^*|x)$, the steering vector $\delta_l$ is inherently constructed to align with the unembedding direction $w_{t^*}$ (i.e., $\delta_l \approx \alpha w_{t^*}$).
\end{itemize}

This mathematically validates our claim in Section~\ref{ssec:editing_abstraction}: ROME's weight update is equivalent to hard-coding a steering vector that is functionally dependent on the oracle label $t^*$.

\section{Localization and Identification of Factual Interference}
\label{app:rome_localization_final}

This section describes the procedure for identifying factual interference within the model's layers. While traditional causal analysis focuses on facilitatory sites, our study prioritizes the extraction of disruptive layers.

\subsection{Causal Tracing and the ROME Reference Site}
The ROME framework employs \textbf{Causal Tracing} to evaluate the contribution of each layer to a specific factual prediction. By calculating the \textbf{Average Indirect Effect (AIE)}, the method quantifies how the probability of a target token $t^*$ recovers when a layer's activation is restored to its ``clean'' state during a corrupted inference run.

In standard practice, ROME identifies a specific layer with the maximum positive AIE to serve as the singular editing site. We denote this traditional reference index as $l_{\text{ROME}}$:
\begin{equation}
    l_{\text{ROME}} = \arg\max_{l \in \{1, \dots, L\}} \text{AIE}_l.
\end{equation}

\subsection{Defining the Interference Set as $l^*$}
Our research shifts the focus toward layers that exhibit a negative causal impact. We define the set of \textbf{Interference Layers} as $l^*$. This set comprises all layers where the restoration of clean activations leads to a further suppression of the target probability:
\begin{equation}
    l^* = \{ l \mid \text{AIE}_l = P_{\text{restored}}(t^*) - P_{\text{corr}}(t^*) < 0 \}.
\end{equation}
A negative AIE suggests that these layers host factual conflicts or inhibitory mechanisms relative to the target fact.

\subsection{Detailed Algorithm}
The following algorithm formalizes the extraction of the interference set $l^*$ alongside the traditional ROME index $l_{\text{ROME}}$.

\begin{algorithm}
\caption{Extraction of Interference Layer Set $l^*$}
\label{alg:extended_causal_tracing}
\begin{algorithmic}[1]
\REQUIRE Model $\mathcal{M}$, prompt $x$ with subject $s$, target $t^*$, noise scale $\sigma$.
\ENSURE $l^*$ and $l_{\text{ROME}}$.
\STATE $P_{\text{clean}}(t^*), \{h_l\}_{l=1}^L \gets \mathcal{M}(x)$ \COMMENT{Baseline Run}
\STATE $x_{\text{corr}} \gets \text{Embed}(x) + \epsilon, \epsilon \sim \mathcal{N}(0, \sigma^2 I)$ \COMMENT{Corrupted Run}
\STATE $P_{\text{corr}}(t^*) \gets \mathcal{M}(x_{\text{corr}})$
\STATE Initialize empty set $l^* = \emptyset$
\FOR{each layer $l \in \{1, \dots, L\}$}
    \STATE $P_{\text{restored}}(t^*) \gets \mathcal{M}(x_{\text{corr}} \mid \text{do}(h_l^{(s)} \leftarrow h_l))$
    \STATE $\text{AIE}_l \gets P_{\text{restored}}(t^*) - P_{\text{corr}}(t^*)$
    \IF{$\text{AIE}_l < 0$}
        \STATE $l^* \gets l^* \cup \{l\}$
    \ENDIF
\ENDFOR
\STATE $l_{\text{ROME}} \gets \arg\max_{l} (\text{AIE}_l)$
\RETURN $l^*, l_{\text{ROME}}$
\end{algorithmic}
\end{algorithm}

\section{Dataset Descriptions}
\label{app:datasets}

This section provides a comprehensive overview of the datasets utilized in our experiments, organized by their primary evaluation tasks. 

\textbf{Data Selection and Scaling.} Following the data processing methodology in AdaCAD (\citet{wang2025adacad}) and considering computational resource constraints, we conduct our evaluation on representative subsets of the original benchmarks. Specifically, for QA and reasoning tasks, we evaluate on: NQ (3,481 samples), HotpotQA (4,690 samples), NQ-Swap (4,000 samples), TabMWP (1,000 samples), TriviaQA (2,000 samples), and SQuAD (3,000 samples). For summarization tasks, we utilize XSum (600 samples), CNN-DailyMail (600 samples), and TofuEval (300 samples).

\subsection{Question Answering and Reasoning}
\begin{description}
    \item[Natural Questions (NQ)] \cite{kwiatkowski2019natural} consists of real search queries from Google. It requires models to find answers within Wikipedia articles, evaluating real-world information retrieval and extraction.
    \item[HotpotQA] \cite{yang2018hotpotqa} focuses on multi-hop reasoning, where the model must aggregate information across multiple documents to arrive at the final answer.
    \item[TriviaQA] \cite{joshi2017triviaqa} includes a large collection of questions authored by trivia enthusiasts. It provides a challenging test for long-form reading comprehension and open-domain QA.
    \item[SQuAD] \cite{rajpurkar2016squad} is a benchmark where the answer to every question is a span of text from the provided reading passage, testing basic reading comprehension.
    \item[TabMWP] \cite{lu2022dynamic} provides mathematical word problems based on tabular data, necessitating both linguistic understanding and numerical reasoning.
    \item[NQ-Swap] \cite{longpre2021entity} is a robustness benchmark that substitutes ground-truth entities with plausible alternatives. It is used to assess how models resolve conflicts between their internal parameters and the provided external context.
\end{description}

\subsection{Summarization and Faithfulness}
\begin{description}
    \item[CNN-DailyMail (DM)] \cite{hermann2015teaching} contains news stories and multi-line summaries. It is widely used to evaluate both extractive and abstractive summarization capabilities.
    \item[XSum] \cite{narayan2018don} provides BBC articles paired with single-sentence summaries. The task is highly abstractive, demanding significant compression and rephrasing.
    \item[TofuEval] \cite{tang2024tofueval} is designed to measure the factual consistency and faithfulness of models, particularly in scenarios involving fictitious information or knowledge unlearning.
\end{description}

\section{Details of Decoding-time Correction Methods}
\label{sec:decoding_methods}
Following the discussion in the main text, this section provides the formal definitions for the decoding-time strategies used to mitigate knowledge conflicts. We denote the context-aware distribution as $P_{ctx} = P(y_t | y_{<t}, x, c)$ and the context-free (null) distribution as $P_{null} = P(y_t | y_{<t}, x)$.
\paragraph{1. CAD \cite{shi2024trusting}}
Context-Aware Decoding (CAD) applies a contrastive objective to the logit space to amplify the influence of the external context $c$ while suppressing parametric priors. The corrected distribution is formulated as:
\begin{equation}
    P_{CAD} \propto \exp \left( (1 + \alpha) \log P_{ctx} - \alpha \log P_{null} \right)
\end{equation}
where $\alpha \geq 0$ is a hyperparameter that determines the strength of the contextual shift.

\paragraph{2. AdaCAD \cite{wang2025adacad}}
AdaCAD improves upon the static nature of CAD by introducing an adaptive weight $\alpha_t$, allowing the model to dynamically regulate the importance of the context at each decoding step $t$:
\begin{equation}
    P_{AdaCAD} \propto \exp \left( \log P_{ctx} + \alpha_t \log \frac{P_{ctx}}{P_{null}} \right)
\end{equation}
By scaling the log-ratio of the two distributions, AdaCAD selectively follows the context only when the discrepancy between $P_{ctx}$ and $P_{null}$ indicates a potential knowledge conflict.

\paragraph{3. COIE \cite{yuan2024discerning}}
The Contextual Information-Entropy (COIE) framework utilizes an entropy-based constraint to discern and resolve conflicts. It measures the uncertainty and information gain through the cross-entropy between the context-aware and prior distributions:
\begin{equation}
    \mathcal{H}(P_{ctx}, P_{null}) = -\sum_{y \in \mathcal{V}} P_{ctx}(y) \log P_{null}(y)
\end{equation}
This entropy constraint serves as a diagnostic tool to determine the reliability of the retrieved context, ensuring the model adaptively balances its internal memory with external evidence based on the information-theoretic properties of the current token.
\section{Implementation Details}
\label{appendix:implementation_details}

For the hyperparameter configurations of the decoding methods used in our experiments, we follow the settings established by prior research to ensure fair comparison. The specific details are as follows:

\begin{itemize}
    \item \textbf{CAD:} Following \citet{shi2024trusting}, we set the scaling factor $\alpha = 1$ for all Question Answering (QA) datasets to emphasize the context. For summarization datasets, $\alpha$ is set to $0.5$ to balance the prior knowledge and the source document.
    
    \item \textbf{COIECD:} In accordance with \citet{yuan2024discerning}, the hyperparameter $\lambda$ is fixed at $0.25$ across all tasks. The entropy constraint parameter $\alpha$ follows the same task-specific settings as CAD: $\alpha = 1$ for QA datasets and $\alpha = 0.5$ for summarization datasets.
    
    \item \textbf{ADACAD:} Unlike the fixed settings above, ADACAD(\citet{wang2025adacad}) employs a dynamic adjustment mechanism for the parameter $\alpha$. The value of $\alpha$ is computed instance-wisely, varying according to the quantified degree of knowledge conflict detected between the model's parametric memory and the provided context.
\end{itemize}
For our proposed method, we maintain $\alpha = 1$ across all datasets, while the values of $k$ and $\lambda$ are tailored to the task requirements. Specifically, for QA tasks, we set $k = 10$, with $\lambda = 0.8$ for reasoning-intensive datasets (e.g., HotpotQA) and $\lambda = 1.0$ for standard QA datasets. For summarization tasks, we adopt $k = 5$ and $\lambda = 0.5$ to strike a balance between context faithfulness and linguistic fluency.

\section{$\protect\alpha$ Rectification}
\label{sec:Alpha impact}
\begin{figure}[t]
    \centering
    \includegraphics[width=0.48\textwidth]{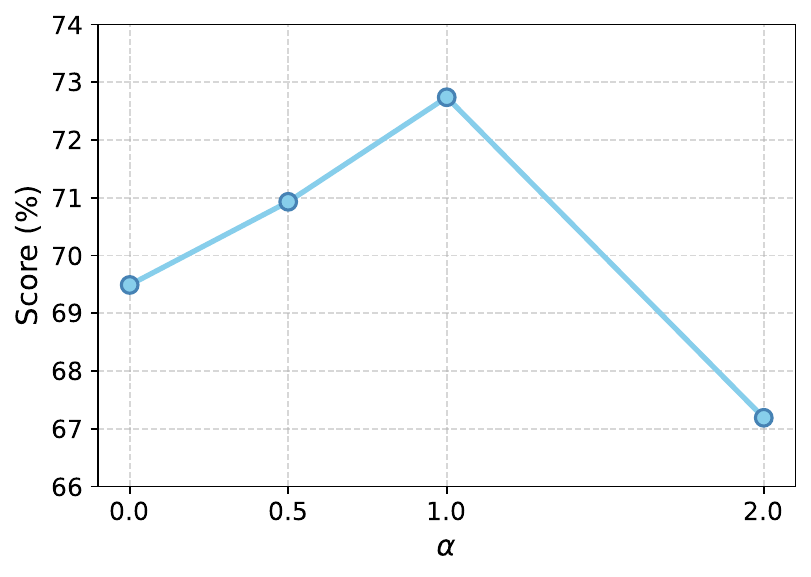}
    \caption{Sensitivity analysis of the scaling factor $\alpha$ on the Natural Questions (NQ) dataset. The peak at $\alpha=1$ indicates that exact neutralization of conflicting signals achieves optimal performance.}
    \label{fig:alpha_sensitivity}
\end{figure}

To further elucidate the mechanism of hidden state intervention, we conduct a sensitivity analysis of the scaling factor $\alpha$ using the Natural Questions (NQ) dataset. This parameter defines a tunable spectrum of intervention intensity: (i) when $0 < \alpha < 1$, the framework performs \textit{partial suppression}, attenuating but not fully neutralizing the conflicting parametric bias; (ii) at $\alpha = 1$, the operation strictly \textit{neutralizes} the suppression, aligning with the pure disinhibition paradigm discussed in Sec.~\ref{ssec:transition}; and (iii) for $\alpha > 1$, the intervention transcends mere removal of inhibition, actively \textit{promoting} the evidence for the target token $\tilde{t}^*$ by flipping the sign of the corrective signal.Experimental results on the NQ dataset, illustrated in Figure~\ref{fig:alpha_sensitivity}, reveal a concave relationship between model performance and the scaling factor. We observe that the performance peaks at $\alpha=1$, confirming that strict neutralization of parametric bias offers the most balanced intervention. While partial suppression ($0 < \alpha < 1$) shows steady improvement over the baseline ($\alpha=0$), aggressive promotion ($\alpha=2$) leads to a performance degradation. This decline suggests that excessive amplification of the corrective signal may introduce over-correction artifacts, potentially distorting the original semantic space of the hidden states.Notably, as $\alpha$ continues to increase to an extreme scale, the intervention will degenerate into a form of direct decoding guided solely by the $\tilde{t}^*$. 
\section{More Case Studies about Direct Decoding and Hidden State Rectification}

\begin{figure}[ht]
    \centering
    \footnotesize 
    \begin{tcolorbox}[
        colback=white, colframe=black, boxrule=0.5pt, arc=0pt, 
        top=2pt, bottom=2pt, left=4pt, right=4pt, boxsep=1pt
    ]
        \textbf{Context:} $\langle$Table$\rangle$ $\langle$Tr$\rangle$ $\langle$Th$\rangle$ Contestant $\langle$/Th$\rangle$ ... $\langle$Td$\rangle$ Amber Brauner $\langle$/Td$\rangle$ $\langle$Td$\rangle$ 37 $\langle$/Td$\rangle$ ... $\langle$Td$\rangle$ Winner March 31 $\langle$/Td$\rangle$ \textit{(truncated)}
        
        \smallskip
        \textbf{Q:} winner of worst cooks in america season 5? 
        
        \textbf{Gold:} Amber Brauner
        
        \textbf{Direct Decoding:} \wrong{Lance Green} 
        
        \textbf{Ours (Rect.):} \correct{Amber Brauner}
    \end{tcolorbox}
    
    \vspace{2pt} 
    \begin{tcolorbox}[
        colback=white, colframe=black, boxrule=0.5pt, arc=0pt, 
        top=2pt, bottom=2pt, left=4pt, right=4pt, boxsep=1pt
    ]
        \textbf{Context:} $\langle$P$\rangle$ The Super Bowl LI Halftime show took place on February 5, 2017... headlined by Lady Gaga, who performed a medley of her songs... \textit{(truncated)}
        
        \smallskip
        \textbf{Q:} who performed the halftime show at super bowl 51? \hfill \textbf{Gold:} Lady Gaga
        
        \textbf{Direct Decoding:} \wrong{liba} Gaga performed the...
        
        \textbf{Ours (Rect.):} \correct{Lady} Gaga performed the...
    \end{tcolorbox}
    
    \vspace{-5pt} 
    \caption{Qualitative examples illustrating the efficacy of our proposed rectification mechanism in both table-based (top) and text-based (bottom) QA tasks.}
    \label{fig:qualitative_cases}
\end{figure}

The qualitative examples in Fig.~\ref{fig:qualitative_cases} illustrate the efficacy and robustness of our proposed rectification mechanism.Unlike naive constrained decoding methods that force the model to output a specific token, our approach operates within the pure "disinhibition" paradigm discussed in Sec.~\ref{ssec:transition}.we do not strictly override the model's generative process. Instead, we steer the underlying representations toward the correct semantic manifold. This distinction is crucial for maintaining model robustness: even if the external knowledge or the target token selection contains slight noise, the latent space retains its internal linguistic coherence.
\end{document}